\documentclass[runningheads]{llncs}
\usepackage[T1]{fontenc}
\usepackage{graphicx}
\usepackage{hyperref}
\begin{document}
\title{Evaluating Large Language Models for Structured Science Summarization in the \\ Open Research Knowledge Graph}
\titlerunning{Evaluating LLMs for Structured Science Summarization}
%
\author{Vladyslav Nechakhin\inst{1}\orcidID{0000-0003-0146-1207} \and
Jennifer D'Souza\inst{2}\orcidID{0000-0002-6616-9509} \and
Steffen Eger\inst{3}\orcidID{0000-0003-4663-8336}}
\authorrunning{V. Nechakhin et al.}
%
\institute{L3S Research Center, Leibniz Univesity Hannover, Hannover, Germany\\
\email{vladyslav.nechakhin@l3s.de} \and
Leibniz Information Centre for Science and Technology, Hannover, Germany\\
\email{jennifer.dsouza@tib.eu} \and
University of Mannheim, Mannheim, Germany\\
\email{steffen.eger@uni-mannheim.de}}
\maketitle              
\begin{abstract}
Structured science summaries or research contributions using properties or dimensions beyond traditional keywords enhances science findability. Current methods, such as those used by the Open Research Knowledge Graph (ORKG), involve manually curating properties to describe research papers' contributions in a structured manner, but this is labor-intensive and inconsistent between the domain expert human curators. We propose using Large Language Models (LLMs) to automatically suggest these properties. However, it's essential to assess the readiness of LLMs like GPT-3.5, Llama 2, and Mistral for this task before application. Our study performs a comprehensive comparative analysis between ORKG's manually curated properties and those generated by the aforementioned state-of-the-art LLMs. We evaluate LLM performance through four unique perspectives: semantic alignment and deviation with ORKG properties, fine-grained properties mapping accuracy, SciNCL embeddings-based cosine similarity, and expert surveys comparing manual annotations with LLM outputs. These evaluations occur within a multidisciplinary science setting. Overall, LLMs show potential as recommendation systems for structuring science, but further finetuning is recommended to improve their alignment with scientific tasks and mimicry of human expertise.

\end{abstract}

\section{Introduction}

The exponential growth of scholarly publications poses a significant challenge for researchers seeking to efficiently explore and navigate the vast landscape of scientific literature \cite{arab2022clustering}. This proliferation of publications necessitates the development of strategies that go beyond traditional keyword-based search methods to facilitate effective and strategic reading practices. In response to this challenge, structured representation of scientific papers has emerged as a valuable approach for enhancing FAIR research discovery and comprehension. By describing research contributions in a structured, machine-actionable format w.r.t. the salient properties of research, also seen as research dimensions, similar such structured papers can be easily compared offering researchers a systematic and quick snapshot of research progress within specific domains, thus enabling them efficient ways to stay updated with research progress.

\begin{figure}[!htb]
\includegraphics[width=\textwidth]{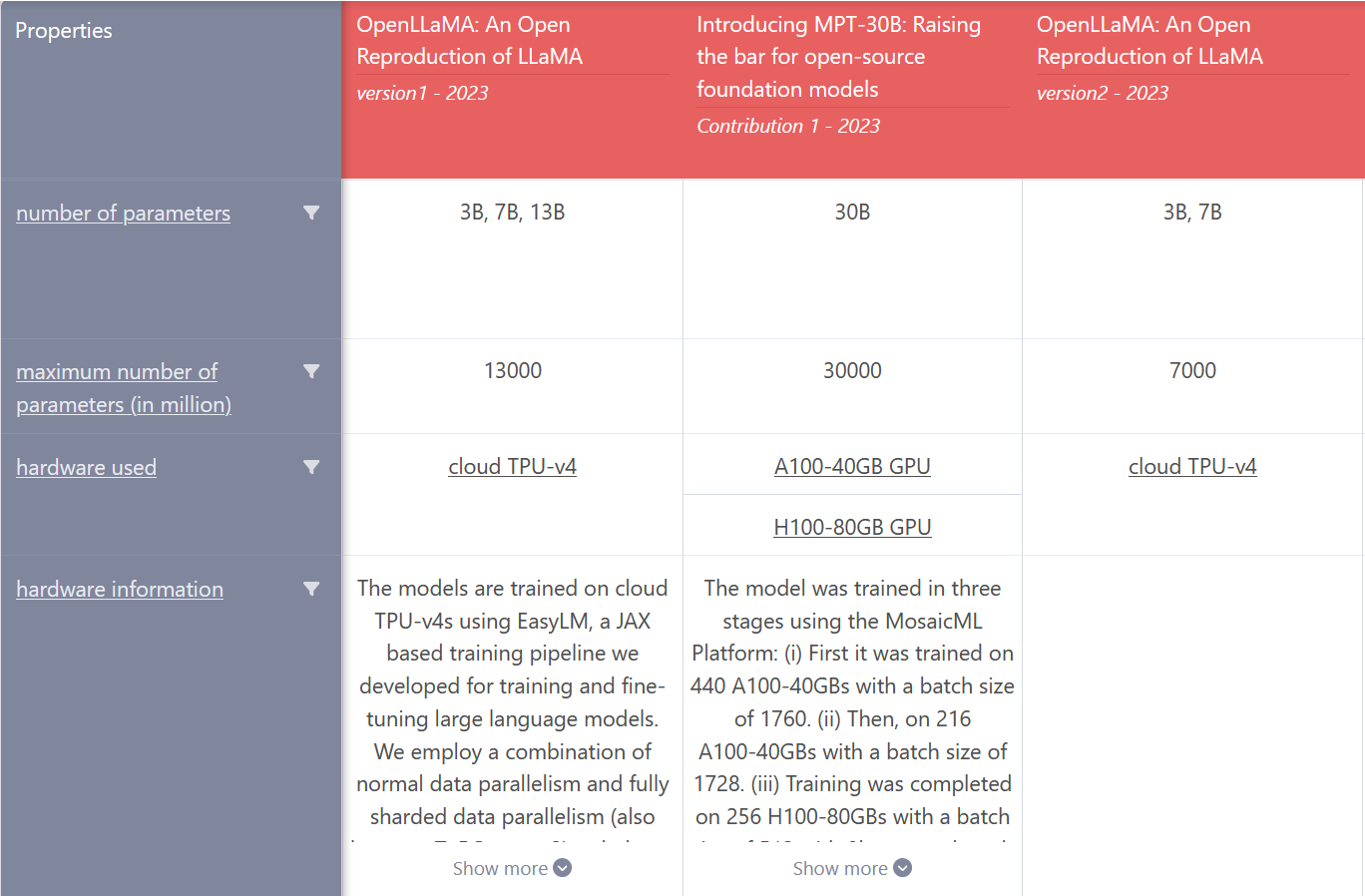}
\caption{ORKG Comparison - A Catalog of Transformer Models (\url{https://orkg.org/comparison/R656113/})\label{fig1-transformers}}
\end{figure}

\begin{figure}[!htb]
\includegraphics[width=\textwidth]{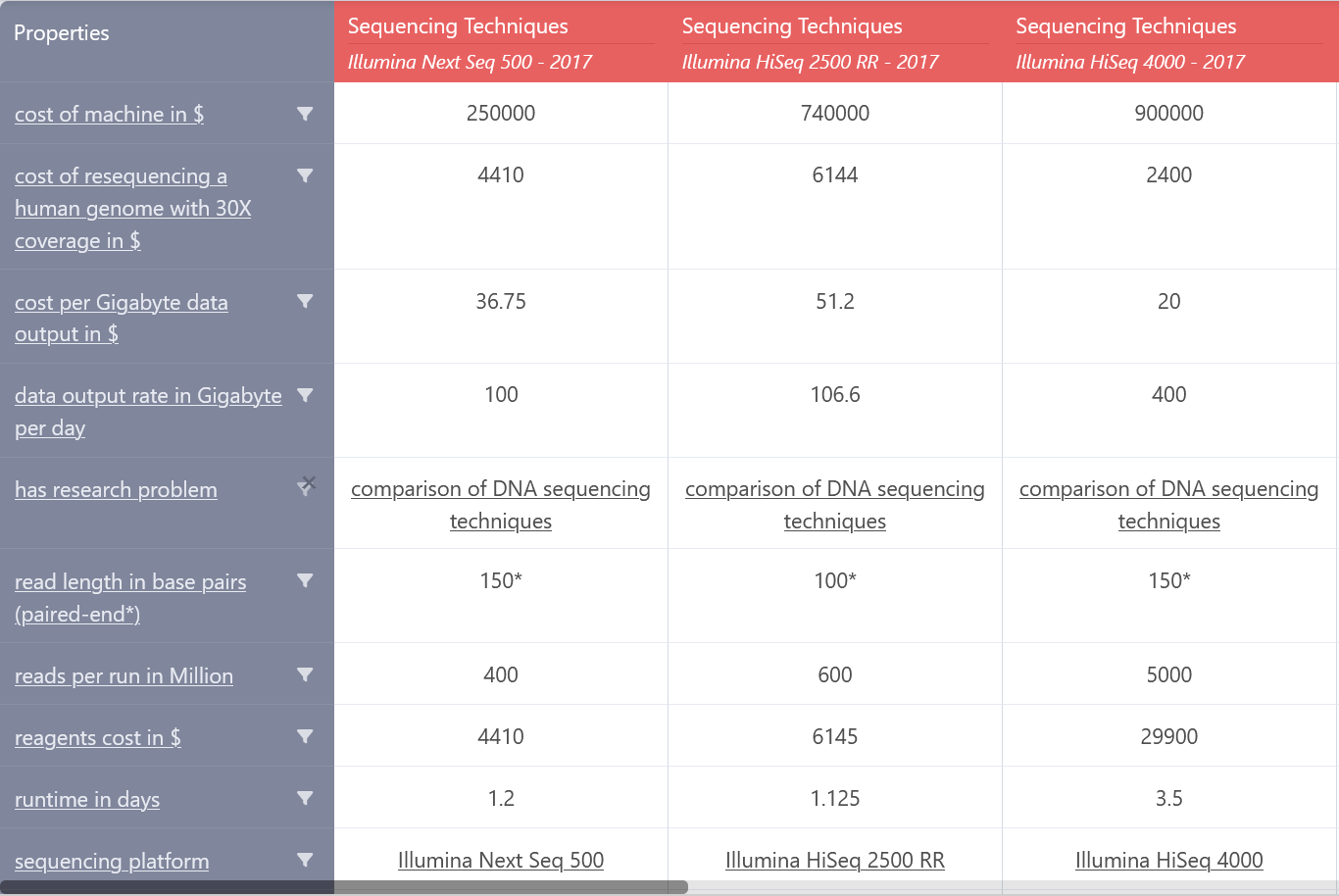}
\caption{ORKG Comparison - Survey of sequencing techniques (\url{https://orkg.org/comparison/R44668/})\label{fig2-dna-sequencing}}
\end{figure}

One notable initiative aimed at publishing structured representations of scientific papers is the Open Research Knowledge Graph (\href{https://orkg.org/}{ORKG}) \cite{auer2020improving}. The ORKG endeavors to describe papers in terms of various research dimensions or properties. For instance, the properties "model family", "pretraining architecture", "number of parameters", "hardware used" that can effectively be applied to offer structured, machine-actionable summaries of research contributions on the research problem "transformer model" in the domain of Computer Science (\autoref{fig1-transformers}). Furthermore, another distinguishing characteristic of the properties is aside from offering a structured summary of a transformer model, they are also generically applicable across various contributions on the same problem thus making the structured paper descriptions comparable. Thus these properties can be explicitly stated as research comparison properties. As another example, papers with the research problem of "DNA sequencing techniques" in the domain of Biology can be described as structured summaries based on the following properties: "sequencing platform", "read length in base pairs", "reagents cost", "runtime in days" (\autoref{fig2-dna-sequencing}). This type of paper description provides a structured framework for understanding and contextualizing research findings. Notably, however, the predominant method in the ORKG for creating structured paper descriptions or research comparisons is manually performed by domain experts. This means that the domain experts based on their prior knowledge and experience on a research problem select and describe the research comparison properties. While this ensures high-quality of the resulting structured papers in the ORKG, the manual annotation cycles cannot effectively scale the ORKG in practice. Specifically, the manual extraction of these salient properties of research or research comparison properties presents two significant challenges: 1) manual annotation a time-consuming process; and 2) it introduces inconsistencies among human annotators, potentially leading to variations in interpretation and annotation.

To address the challenges associated with the manual annotation of research comparison properties, this study tests the feasibility of using pretrained Large Language Models (LLMs) to automatically suggest or recommend research dimensions as candidate properties as a viable alternative solution. Specifically, three different LLM variants, viz. GPT-3.5-turbo \cite{achiam2023gpt}, Llama 2 \cite{touvron2023llama}, and Mistral \cite{jiang2023mistral}, are tested and empirically compared for their advanced natural language processing (NLP) capabilities when applied to the task of recommending research dimensions as candidate properites. Our choice to apply LLMs is based on the following experimental consideration. The multidisciplinary nature of scientific research poses unique challenges to the identification and extraction of salient properties across domains. LLMs, with their ability to contextualize and understand natural language at scale \cite{harnad2024language,karanikolas2023large}, are particularly well-suited to navigate the complexities of interdisciplinary research and recommend relevant dimensions that capture the essence of diverse scholarly works. By automating the extraction process, LLMs aim to alleviate the time constraints associated with manual annotation and ensure a higher level of consistency in the specification of research dimensions by using the same system prompt or fine-tuning on gold-standard ORKG data to better align them to the task. The role of LLMs in this context is to assist domain-expert human annotators rather than replace them entirely. By leveraging the capabilities of LLMs, researchers can streamline the process of dimension extraction and enhance the efficiency and reliability of comparative analysis across diverse research fields. 

In this context, the central research question (RQ) of this study is to examine the performance of state-of-the-art Large Language Models (LLMs) in recommending research dimensions. To address this RQ, we compiled a multidisciplinary, gold-standard dataset of human-annotated scientific papers from the Open Research Knowledge Graph (ORKG), detailed in the Materials and Methods section (see \autoref{sec:dataset}). This dataset includes structured summary property annotations by domain experts. We conducted a detailed comparative evaluation of the domain-expert annotated properties from the ORKG against the dimensions generated by LLMs for the same papers. Our evaluations are based on four unique perspectives: 1) semantic alignment and deviation assessment by GPT-3.5 between ORKG properties and LLM-generated dimensions, 2) fine-grained property mapping accuracy by GPT-3.5, 3) SciNCL \cite{ostendorff2022neighborhood} embeddings-based cosine similarity between ORKG properties and LLM-generated dimensions, and 4) a survey with human experts comparing their annotations of ORKG properties with the LLM-generated dimensions.

As such, the contribution of this work is a comprehensive set of insights into the readiness of LLMs to support human annotators in the task of structuring their research contributions. Our findings reveal a moderate alignment between LLM-generated dimensions and manually annotated ORKG properties, indicating the potential for LLMs to learn from human-annotated data. However, there is a noticeable gap in the mapping of dimensions generated by LLMs and those annotated by domain experts, highlighting the need for fine-tuning LLMs on domain-specific datasets to reduce this disparity. Despite this gap, LLMs demonstrate the ability to capture the semantic relationships between LLM-generated dimensions and ORKG properties, as evidenced by strong correlation results of embeddings similarity. In the survey, the human experts noted that while they were not ready to change their existing property annotations based on the LLM generated dimensions, they highlighted the utility of the auto-LLM recommendation service at the time of creating the structured summary descriptions. This directly informs a future research direction in making LLMs fit for structured science summarization.

\section{Related Work}

The utilization of Large Language Models (LLMs) for various NLP tasks has has seen widespread adoption in recent years \cite{radford2018improving,brown2020language}. Within the realm of scientific literature analysis, researchers have explored the potential of LLMs for tasks such as generating summaries and abstracts of research papers \cite{cai2024sciassess,jin2024comprehensive}, extracting insights and identifying patterns \cite{liang2023can}, aiding in literature reviews \cite{antu2023using}, enhancing knowledge integration  \cite{latif2023knowledge}, etc. However, the specific application of LLMs for recommending research dimensions to obtain structured representations of research contributions is a relatively new area of investigation that we explore in this work. Furthermore, to offer insights into the readiness of LLMs over our novel task, we perform a comprehensive set of evaluations comparing the LLM-generated research dimensions and the human expert annotated properties. As a straightforward preliminary evaluation, we measure the semantic similarity between the LLM and human annotated properties. To do this, we employ a specialized language model tuned for the scientific domain to create embeddings for the respective properties.

The development of domain-specific language models has been a significant advancement in NLP. In the scientific domain, a series of specialized models have emerged. SciBERT, introduced by Beltagy et al. \cite{beltagy2019scibert}, was the first language model tailored for scientific texts. This was followed by SPECTER, developed by Cohan et al. \cite{cohan2020specter}. More recently, Ostendorff et al. introduced SciNCL \cite{ostendorff2022neighborhood}, a language model designed to capture semantic similarity between scientific concepts by leveraging pre-trained BERT embeddings. SciNCL has demonstrated its effectiveness in evaluating the nuances of scientific concepts, making it an ideal choice for assessing LLM-generated dimensions in scientific literature analysis. In this study, we utilize SciNCL, the most recent and advanced variant, to generate embeddings for ORKG properties and LLM-generated dimensions.

In the context of evaluating LLM-generated dimensions against manually curated properties, several studies have employed similarity measures to quantify the relatedness between the two sets of textual data. One widely used metric is cosine similarity, which measures the cosine of the angle between two vectors representing the dimensions \cite{singhal2001modern}. This measure has been employed in various studies, such as Yasunaga et al. \cite{yasunaga2019scisummnet}, who used cosine similarity to assess the similarity between automatically generated summaries by LLMs and human-written annotations. Similarly, Banerjee et al. \cite{banerjee2023benchmarking} employed cosine similarity as a metric to benchmark the accuracy of LLM-generated answers of autonomous conversational agents. In contrast to cosine similarity, other studies have explored alternative similarity measures for evaluating LLM-generated content. For instance, Jaccard similarity measures the intersection over the union of two sets, providing a measure of overlap between them \cite{verma2020comparative}. This measure has been employed in tasks such as document clustering and topic modeling \cite{ferdous2009efficient,o2015analysis}. Jaccard similarity offers a distinct perspective on the overlap between manually curated and LLM-generated properties, as it focuses on the shared elements between the two sets rather than their overall similarity. We considered both cosine and Jaccard similarity in our evaluation, however, based on our embedding representation, we ultimately chose to use cosine similarity as our distance measure.

Finally, aside from the straightforward similarity computations between the two sets of properties, we also leverage the capabilities of LLMs as evaluators. The utilization of LLMs as evaluators in various NLP tasks has been proven to be a successful approach in a number of recent publications. For instance, Kocmi and Federmann \cite{kocmi2023large} demonstrated the effectiveness of GPT-based metrics for assessing translation quality, achieving state-of-the-art accuracy in both reference-based and reference-free modes. Similarly, the Eval4NLP 2023 shared task, organized by Leiter et al. \cite{leiter2023eval4nlp}, explored the use of LLMs as explainable metrics for machine translation and summary evaluation, showcasing the potential of prompting and score extraction techniques to achieve results on par with or even surpassing recent reference-free metrics. In our study, we employ the GPT-3.5 model as an evaluator, leveraging its capabilities to assess the quality of LLM-generated research dimensions.

In summary, previous research has laid the groundwork for evaluating LLMs' performance in scientific literature analysis, and our study builds upon these efforts by exploring the application of LLMs for recommending research dimensions and evaluating their quality using specialized language models and similarity measures.

\section{Materials and Methods}

This section is organized into three subsections. In the first subsection, the creation of the gold-standard evaluation dataset from the ORKG with domain-expert, human-annotated research comparison properties used for assessing the similarity with the LLM generated properties is described. The second subsection provides an overview of the three LLMs, viz. GPT-3.5, Llama 2, and Mistral, applied to automatically generate the research comparison properties, highlighting their respective technical characteristics. Lastly, the third subsection discusses the various evaluation methods used in this study offering differing perspectives on the similarity comparison of the ORKG properties for the instances in our gold-standard dataset versus those generated by the LLMs.

\subsection{Material: Our Evaluation Dataset}
\label{sec:dataset}

\begin{figure}[!htb]
\centering
\includegraphics[width=10.5 cm]{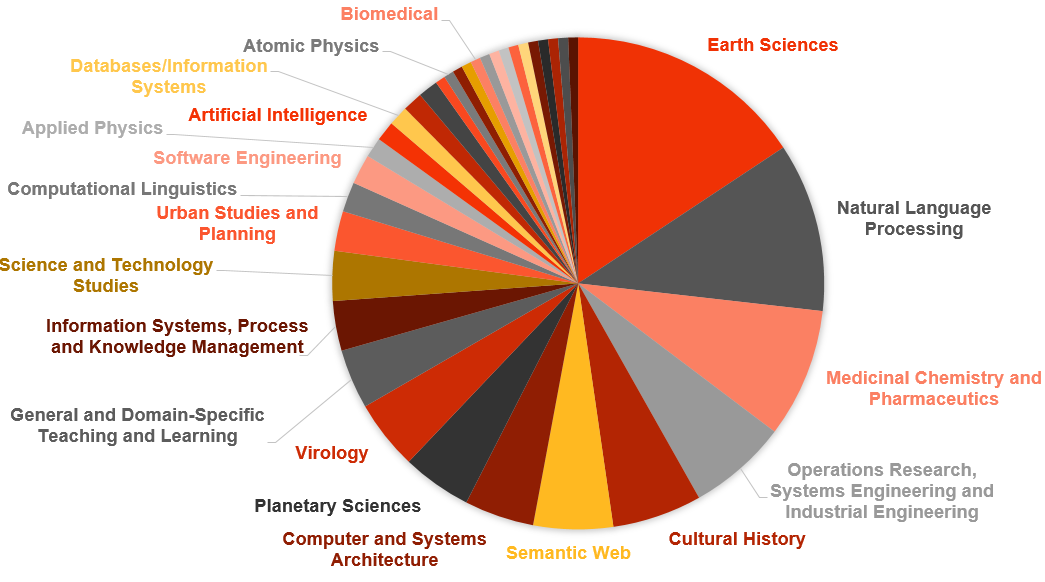}
\caption{Research field distribution of the selected papers in our evaluation dataset containing domain-expert human-annotated properties that were applied to represent the paper's structured contribution descriptions in the Open Research Knowledge Graph (ORKG). \label{RF distribution}}
\end{figure}

As alluded to in the Introduction, a central \textbf{RQ} of this work is to compare the research dimensions generated by three different LLMs with the human-annotated research comparison properties in the ORKG. For this, we created an evaluation dataset of annotated research dimensions based on the ORKG. As a starting point, we curated a selection of \href{https://orkg.org/comparisons}{ORKG Comparisons} by selecting comparisons that were created by experienced ORKG users. These users had varied research backgrounds. The selection criteria of comparisons from these users were as follows: the comparisons had to have at least 3 properties and contain at least 5 contributions, since we wanted to ensure that the properties were not too sparse representation of a research problem, but were those that generically reflected a research comparison over several works. On application of this criteria, the resulting dataset comprised 103 ORKG Comparisons. These selected gold-standard comparisons contained 1,317 papers from 35 different research fields addressing over 150 distinct research problems. The gold-standard dataset can be downloaded from the \href{ https://doi.org/10.25835/6oyn9d1n}{Leibniz University Data Repository}. The selection of comparisons ensured the diversity of research fields’ distribution, containing Earth Sciences, Natural Language Processing, Medicinal Chemistry and Pharmaceutics, Operations Research, Systems Engineering, Cultural History, Semantic Web and others. See \autoref{RF distribution} for the full distribution of research fields in our dataset. 

Once we had the comparisons, we then looked at the individual structured papers within each comparison and extracted their human annotated properties. Thus our resulting dataset, is highly multidisciplinary, comprising structured paper instances from the ORKG with their corresponding domain expert property annotations across different fields of research. For instance, the properties listed below were extracted form the comparison "A Catalog of Transformer Models" (\autoref{fig1-transformers}): 

\begin{verbatim}
["has model", "model family", "date created", "organization",
"innovation", "pretraining architecture", "pretraining task",
"fine-tuning task", "training corpus", "optimizer", 
"tokenization", "number of parameters", "maximum number of 
parameters (in million)", "hardware used", "hardware 
information", "extension", "has source code", "blog post", 
"license", "research problem"]
\end{verbatim}

Another example of structured paper's properties in the comparison "Survey of sequencing techniques" (\autoref{fig2-dna-sequencing}) is as follows: 

\begin{verbatim}
["cost of machine in $", "cost of resequencing a human genome 
with 30X coverage in $", "cost per Gigabyte data output in $", 
"data output rate in Gigabyte per day", "has research problem", 
"read length in base pairs (paired-end*) ", "reads per run in 
Million", "reagents cost in $", "runtime in days", "sequencing 
platform", "total data output yield in Gigabyte"]
\end{verbatim}

The aforementioned dataset is now the gold-standard that we use in the evaluations for the LLM-generated research dimensions. In this section, we provide a clear distinction between the terminology of ORKG \textit{properties} and LLM-generated \textit{research dimensions}. According to our hypothesis, ORKG properties are not necessarily identical to research dimensions. Contribution properties within ORKG relate to specific attributes or characteristics associated with individual research papers in a comparison, outlining aspects such as authorship, publication year, methodology, and findings. Conversely, research dimensions encapsulate the multifaceted aspects of a given research problem, constituting the nuanced themes or axes along which scholarly investigations are conducted. ORKG contribution properties offer insights into the attributes of individual papers, research dimensions operate at a broader level, revealing the finer-grained thematic fundamentals of research endeavors. While ORKG contribution properties focus on the specifics of research findings, research dimensions offer a more comprehensive context for analyzing a research question that can be used for finding similar papers that share the same dimensions. In order to test the alignment of LLM-generated research dimensions to ORKG properties, several LLMs were selected to be compared, as described in the next section.

\subsection{Method: Three Large Language Models Applied for Research Dimensions Generation}

In this section, we discuss the LLMs applied for automated research dimensions generation as well as the task-specific prompt that was designed as input to the LLM.

\subsubsection{The three LLMs}

To test the automated generation of research dimensions, we tested and compared the output from three different state-of-the-art LLMs with comparable parameter counts, namely GPT-3.5, Llama 2 and Mistral.

GPT-3.5, developed by OpenAI, is one of the most influential LLMs to date, its number of parameters is not publicly disclosed \cite{ChatGPT}. 
In comparison, it's predecessor GPT-3 models come in different sizes and contain from 125 million parameters for the smallest model to 175 billion for the largest \cite{brown2020language}. GPT-3.5 has demonstrated exceptional performance on a range of NLP tasks including translation, question answering and text completion. Notably, the capabilities of this model are accessed through the OpenAI API, since the model is closed-source, which limits direct access to its architecture for further exploration or customization.

Llama 2 by Meta AI \cite{touvron2023llama}, the second iteration of the Llama LLM, represents a significant advancement in the field. Featuring twice the context length of its predecessor Llama 1 \cite{touvron2023llama1}, Llama 2 offers researchers and practitioners a versatile tool for working with NLP. Importantly, Llama 2 is available free of charge for both research and commercial purposes, with multiple parameter configurations available, including a 13 billion parameter option. In addition, the model supports fine-tuning and self-hosting, which enhances its adaptability to a variety of use cases.

Mistral, developed by Mistral AI, is a significant competitor in the landscape of LLMs. With a parameter count of 7.3 billion, Mistral demonstrates competitive performance in various benchmarks, often outperforming Llama 2 despite its smaller size \cite{jiang2023mistral}. In addition, Mistral is distinguished by its open source code released under the Apache 2.0 license, making it easily accessible for research and commercial applications. Notably, Mistral has an 8k context window, compared to 4k context window of Llama 2, which allows for a more complete understanding of context \cite{thakkar2023comprehensive}.

Overall, GPT-3.5, despite its closed source nature, remains influential in NLP research, with a vast number of parameters that facilitate its generic task performance in the context of a wide variety of applications. Conversely, Llama 2 and Mistral, with their open source nature, provide flexibility and accessibility for researchers and developers while displaying similar performance characteristics to GPT \cite{llmLeaderboard}. Released shortly after Llama 2, Mistral, in particular, shows notable performance advantages over Llama 2, highlighting the rapid pace of innovation and improvement in the development of LLMs. These differences between the models lay the groundwork for assessing their alignment with manually curated properties from ORKG and determining their potential for automated research metadata creation and retrieval of related work.

\begin{table}[!htb] 
\caption{Prompt variations utilizing different prompt engineering techniques to instruct LLMs for the research dimensions generation task.\label{table prompts}}
\begin{tabular}{|p{1.8cm}|p{7cm}|p{3.1cm}|}
\hline
\textbf{Prompting technique}	& \textbf{System prompt}	& \textbf{Output example}\\
\hline
Zero-Shot & 
You will be provided with a research problem, your task is to list dimensions that are relevant to find similar papers for the research problem. Respond only in the format of a python list. & 

["Natural Language Processing", "Text Analysis", "Machine Learning", "Deep Learning", "Information Retrieval", "Artificial Intelligence", "Language Models", "Document Summarization"] \\ \hline

Few-Shot & 
You will be provided with a research problem, your task is to list dimensions that are relevant to find similar papers for the research problem. Respond only in the format of a python list. \newline 
The following are two successfully completed task examples. \newline
Research problem: "Transformer models" \newline
Research dimensions: ['model', 'date created', 'pretraining architecture', 'pretraining task', 'training corpus', 'optimizer', 'tokenization', 'number of parameters', 'license'] \newline
Research problem: "Liposomes as drug carriers" \newline
Research dimensions: ['Type of nanocarrier', 'Nanoparticle preparation method', 'Lipid composition', 'Drug type', 'Particle size'] & 

['Summarization approach', 'Document type', 'Language', 'Evaluation metric', 'Model type', 'Training dataset', 'Compression ratio', 'Summary length'] \\ \hline

Chain-of-Thought & 
You will be provided with a research problem, your task is to list dimensions that are relevant to find similar papers for the research problem. Provide justification why each dimension is relevant for finding similar papers. Think step-by-step. At the end combine all the relevant dimensions in the format of a python list. & 

["Task/Methodology", "Domain/Genre", "Evaluation Metrics", "Language", "Input/Output Format", "Deep Learning/Traditional Methods", "Applications"] \\

\hline
\end{tabular}
\end{table}

\subsubsection{Research dimensions generation prompt for the LLMs}

LLM’s performance on a particular task is highly dependent on the quality of the prompt. To find the optimal prompt methodology, our study explores various established prompting techniques, including zero-shot \cite{radford2019language}, few-shot  \cite{brown2020language}, and chain-of-thought prompting \cite{wei2022chain}. The simplest type of prompt is a zero-shot approach, wherein pretrained LLMs owing to the large-scale coverage of human tasks within their pretraining datasets demonstrate competence in task execution without prior exposure to specific examples. While zero-shot prompting provides satisfactory results for certain tasks, some of the more complex tasks require few-shot prompting. By providing the model with several examples, few-shot prompting enables in-context learning, potentially enhancing model performance. Another popular technique is chain-of-thought prompting, which instructs the model to think step-by-step. Guiding the LLM through sequential steps helps to break down complex tasks into manageable components that are easier for the LLM to complete. 

In this study, the task involves providing the LLM with a research problem, and based on this input, the model should suggest research dimensions that it finds relevant to structure contributions from similar papers that address the research problem. Our system prompts for each technique along with the examples of output for a research problem "Automatic text summarization" from GPT-3.5 are shown in the \autoref{table prompts}. 

Our analysis shows that the utilization of more advanced prompting techniques did not necessarily result in superior outcomes, which leads us to believe that our original zero-shot prompt is sufficient for our task completion. The absence of discernible performance improvements with the adoption of more complex prompting techniques highlights the effectiveness of the initial zero-shot prompt in aligning with the objectives of research dimension extraction. Consequently, we will be apply the zero-shot prompt methodology.

To test the alignment of LLM-generated research dimensions to ORKG properties, each of the LLMs was given the same prompt to create a list of dimensions that are relevant to find similar papers based on the provided research problem. \autoref{table dimensions} shows the comparison between some of the manually created properties form ORKG and the research dimensions, provided by GPT-3.5, Llama 2 and Mistral. 

\begin{table}[!htb] 
\caption{Comparison of manually created ORKG properties with LLM-generated research dimensions for the same papers.\label{table dimensions}}
\begin{tabular}{|p{1.5cm}|p{2.5cm}|p{2.0cm}|p{3cm}|p{3cm}|}
\hline
\textbf{Research problem}	& \textbf{ORKG properties}	& \textbf{GPT-3.5 dimensions}    & \textbf{Llama 2 dimensions}	& \textbf{Mistral dimensions}\\
\hline
\href{https://orkg.org/comparison/R141123/}{Limestone mineral identification using Hyperion imagery} &
'Data used', 
'Research problem', 
'Study Area', 
'Software', 
'Analysis', 
'Processing', 
'Field instrument', 
'Reference' &
'Mineralogy', 
'Limestone', 
'Hyperion imagery', 
'Identification' &
'Remote Sensing', 
'Hyperspectral Imaging', 
'Mineral Identification', 
'Limestone', 
'Hyperion Imagery', 
'Geological Mapping', 
'Image Classification', 
'Spectral Analysis', 
'Earth Observation', 
'Geoscience' &
'Limestone',
'Mineral identification',
'Remote sensing',
'Hyperion imagery',
'Spectral signatures',
'Band ratios',
'Textural features',
'Shape descriptors',
'Machine learning algorithms',
'Support Vector Machines',
'Decision trees',
'Random forests' \\

\hline
\href{https://orkg.org/comparison/R204005/}{Do human noroviruses infect animals} &
'Material', 
'Method', 
'Symptoms and signs', 
'Research problem', 
'Species', 
'Location', 
'Antigen used', 
'Immunoglobulin class', 
'Primers', 
'Noroviruses found', 
'Norovirus genotype' &
'viral infectivity', 
'host range', 
'cross-species transmission', 
'zoonotic potential' &
'Host species',
'Viral genotype',
'Transmission route',
'Infection prevalence',
'Pathogenesis',
'Diagnostic methods',
'Epidemiology',
'Evolutionary history',
'Interaction with other pathogens',
'Public health implications' &
'Norovirus',
'Animals',
'Transmission',
'Species',
'Symptoms',
'Epidemiology',
'Serotypes',
'Immunity',
'Detection methods',
'Genome analysis' \\
\hline
\end{tabular}
\end{table}

\subsection{Method: Three Types of Similarity Evaluations between ORKG Properties vs. LLM-generated Research Dimensions}

This section outlines the methodology used to evaluate our dataset, namely the automated evaluation of semantic alignment and deviation as well as mapping between ORKG properties and LLM-generated research dimensions performed by GPT-3.5. Additionally, we present our approach to calculating embedding similarity between properties and research dimensions.

\subsubsection{Semantic alignment and deviation evaluations using GPT-3.5}

To measure the semantic similarity between between ORKG properties and LLM-generated research dimensions, we conducted semantic alignment and deviation assessments using an LLM-based evaluator. In this context, semantic alignment refers to the degree to which two sets of concepts share similar meanings, whereas semantic deviation assesses how far apart they are in terms of meaning. As the LLM evaluator, we leveraged GPT-3.5. As input it was provided with both the lists of properties from ORKG and the dimensions extracted by the LLMs, in a string format, per research problem. Semantic alignment was rated on a scale from 1 to 5, using the following system prompt to perform this task:

\begin{verbatim}
You will be provided with two lists of strings, your task is to
rate the semantic alignment between the lists on the scale form
1 to 5.Your response must only include an integer representing 
your assessment of the semantic alignment, include no other 
text.
\end{verbatim}

Additionally, the prompt included the detailed description of the scoring system shown in the \autoref{table alignment}.

\begin{table}[!htb] 
\caption{Description of the semantic alignment scores provided in the GPT-3.5 system prompt.\label{table alignment}}
\begin{tabular}{|l|p{9cm}|}
\hline
\textbf{Score}	& \textbf{Description}\\
\hline
1 - Strongly Disaligned &
The strings in the lists have minimal or no semantic similarity. \\
2 - Disaligned &
The strings in the lists have limited semantic alignment. \\
3 - Neutral &
The semantic alignment between the lists is moderate or average. \\
4 - Aligned &
The strings in the lists show substantial semantic alignment. \\
5 - Strongly Aligned &
The strings in the lists exhibit high semantic coherence and alignment. \\
\hline
\end{tabular}
\end{table}

To further validate the accuracy of our alignment scores, we leveraged GPT-3.5 as an evaluator again but this time to generate the semantic deviation scores. By using this contrastive alignment versus deviation evaluation method, we can cross-reference where the LLM evaluator displays a strong agreement in its evaluation and assess the evaluations for reliability. Specifically, we evaluate the same set of manually curated properties and LLM-generated research dimensions using both agents, with the expectation that the ratings will exhibit an inverse relationship. That is, high alignment scores should correspond to low deviation scores, and vice versa. The convergence of these opposing measures would provide strong evidence for the validity of our evaluation results. Similar to the task of alignment rating, the system prompt below was used to instruct GPT-3.5 to measure semantic deviation where the ratings described in \autoref{table deviation} was also part of the prompt.

\begin{verbatim}
You will be provided with two lists of strings, your task is to
rate the semantic deviation between the lists on the scale form
1 to 5. Your response must only include an integer representing
your assessment of the semantic deviation, include no other 
text.
\end{verbatim}

\begin{table}[!htb] 
\caption{Description of the semantic deviation scores provided in the GPT-3.5 system prompt.\label{table deviation}}
\begin{tabular}{|l|p{8cm}|}
\hline
\textbf{Score}	& \textbf{Description}\\
\hline
1 - Minimal Deviation &
The strings in the lists show little or no semantic difference. \\
2 - Low Deviation &
The semantic variance between the lists is limited. \\
3 - Moderate Deviation &
There is a moderate level of semantic difference between the strings in the lists. \\
4 - Substantial Deviation &
The lists exhibit a considerable semantic gap or difference. \\
5 - Significant Deviation &
The semantic disparity between the lists is pronounced and substantial. \\
\hline
\end{tabular}
\end{table}

By combining these two evaluations, we can gain a more nuanced understanding of the relationship between the ORKG properties and LLM-generated research dimensions.

\subsubsection{ORKG property and LLM-generated research dimension mappings by GPT-3.5}

To further analyse the relationships between ORKG properties and LLM-generated research dimensions, we used GPT-3.5 to find the mappings between individual properties and dimensions. This approach diverges from the previous semantic alignment and deviation evaluations, which considered the lists as a whole. Instead, we instructed GPT-3.5 to identify the number of properties that exhibit similarity with individual research dimensions. This was achieved by providing the model with the two lists of properties and dimensions and prompting it to count the number of similar values between the lists. 

The system prompt used for this task was as follows:

\begin{verbatim}
You will be provided with two lists of strings, your task is to 
count how many values from list1 are similar to values of 
list2. Respond only with an integer, include no other text.
\end{verbatim}

By leveraging GPT-3.5's capabilities in this manner, we were able to count the number of LLM-generated research dimensions that are related to individual ORKG properties. The mapping count provides a more fine-grained insight into the relationships between the properties and dimensions.

\subsubsection{Scientific embeddings-based semantic distance evaluations}

To further examine the semantic relationships between ORKG properties and LLM-generated research dimensions, we employed embeddings-based approach. Specifically, we utilized SciNCL to generate vector embeddings for both the ORKG properties and the LLM-generated research dimensions. These embeddings were then compared using cosine similarity as a measure of semantic similarity. We evaluated the similarity of ORKG properties to the research dimensions generated by GPT-3.5, Llama 2, and Mistral. Additionally, we compared the LLM-generated dimensions to each other, providing insights into the consistency and variability of the research dimensions generated by different LLMs. By leveraging embeddings-based evaluations, we were able to quantify the semantic similarity between the ORKG properties and the LLM-generated research dimensions, as well as among the dimensions themselves. 

\subsubsection{Human assessment survey comparing ORKG properties with LLM-generated research dimensions}

We conducted a survey to evaluate the utility of LLM generated dimensions in the context of domain-expert annotated ORKG properties. The survey was designed to solicit the impressions of domain experts when shown their original annotated properties versus the research dimensions generated by GPT-3.5. We selected participants who are experienced at creating structured paper descriptions in the ORKG. These participants included \href{https://orkg.org/about/28/Curation_Grants}{ORKG curation grant} participants, ORKG employees, and authors whose comparisons were displayed on the ORKG \href{https://orkg.org/featured-comparisons}{featured comparisons} page. Each participant was given a maximum of 5 surveys, for five different papers they structured respectively, each evaluating properties versus research dimensions. They had the choice to respond to one, some, or all of them. At the end, we received 23 total responses to our survey corresponding to 23 different papers.

\begin{figure}[!htb]
\includegraphics[width=\textwidth]{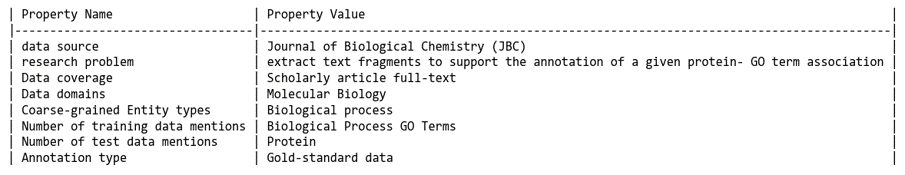}
\caption{Example of ORKG properties shown to survey respondents\label{figure stage1 orkg}}
\end{figure}

\begin{figure}[!htb]
\includegraphics[width=\textwidth]{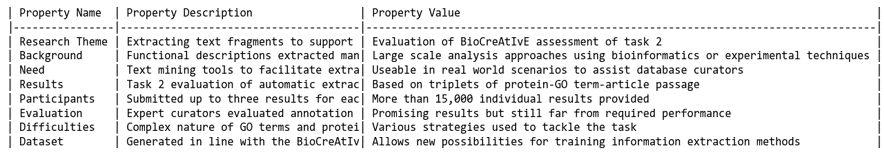}
\caption{Example of GPT dimensions shown to survey respondents\label{figure stage1 gpt}}
\end{figure}

The survey itself consisted of five questions, most of which were designed on a Likert scale, to guage the domain expert assessment of the effectiveness of the LLM-generated research dimensions. Per survey, as evaluation data, participants were presented with two tables: one including their annotated ORKG property names and values (\autoref{figure stage1 orkg}), and the second one consisting of research dimension name, its description, and value generated by GPT-3.5 from a title and abstract of the same paper (\autoref{figure stage1 gpt}). Following this data was the survey questionnaire. The questions asked participants to rate the relevance of LLM-generated research dimensions, consider making edits to the original ORKG structured contribution, and evaluate the usefulness of LLM-generated content as suggestions before creating their structured contributions. Additionally, participants were asked to describe how LLM-generated content would have been helpful and rate the alignment of LLM-generated research dimensions with the original ORKG structured contribution. The survey questionnaire is shown below:

\begin{enumerate}
    \item How many of the properties generated by ChatGPT are relevant to your ORKG structured contribution? (Your answer should be a number)
    \item Considering the ChatGPT-generated content, would you consider making edits to your original ORKG structured contribution?
    \item If the ChatGPT-generated content were available to you as suggestions before you created your structured contribution, would it have been helpful?
    \begin{enumerate}
        \item If you answered "Yes" to the question above, could you describe how it would have been helpful?
    \end{enumerate}
    \item On a scale of 1 to 5, please rate how well the ChatGPT-generated response aligns with your ORKG structured contribution.
    \item We plan to release an AI-powered feature to support users in creating their ORKG contributions with automated suggestions. In this context, please share any additional comments or thoughts you have regarding the given ChatGPT-generated structured contribution and its relevance to your ORKG contribution.
\end{enumerate}

The subsequent section will present the results obtained from these methodologies, providing insights into the similarity between ORKG properties and LLM-generated research dimensions.

\section{Results and Discussion}

This section presents the results of our evaluation, which aimed to assess the LLMs' performance on the task of recommending research dimensions by calculating the similarity between ORKG properties and LLM-generated research dimensions. We employed three types of similarity evaluations: semantic alignment and deviation assessments, property and research dimension mappings using GPT-3.5, and embeddings-based evaluations using SciNCL.

\subsection{Semantic alignment and deviation evaluations}

\begin{figure}[!htb]
\includegraphics[width=\textwidth]{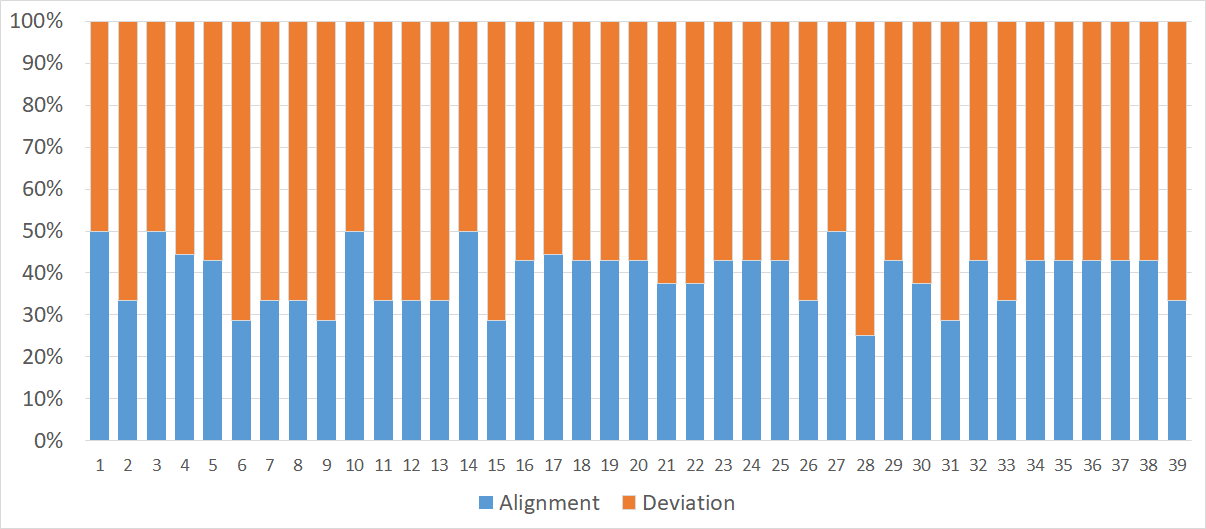}
\caption{Proportional values of semantic alignment and deviation between ORKG properties and GPT-generated dimensions\label{figure Alignment Deviation}}
\end{figure}

The average alignment between paper properties and research field dimensions was found to be 2.9 out of 5, indicating a moderate level of alignment. In contrast, the average deviation was 4.1 out of 5, suggesting a higher degree of deviation. When normalized, the proportional values translate to 41.2\% alignment and 58.8\% deviation (\ref{figure Alignment Deviation}). These results imply that while there is some alignment between paper properties and research field dimensions, there is also a significant amount of deviation, highlighting the difference between the concepts of structured paper's properties and research dimensions. This outcome supports our hypothesis that LLM-based research dimensions generated solely from a research problem, relying on LLM-encoded knowledge, may not fully capture the nuanced inclinations of domain experts when they annotate ORKG properties to structure their contributions, where the domain experts have the full paper at hand. We posit that an LLM, not explicitly tuned on the scientific domain, despite its vast parameter space, is not able to emulate human expert subjectivity to structure contributions.

\subsection{Property and research dimension mappings}

We examine our earlier posited claim in a more fine-grained manner by comparing the property versus research dimension mappings. The average number of mappings was found to be 0.33, indicating a low number of mappings between paper properties and research field dimensions. The average ORKG property count was 4.73, while the average GPT dimension count was 8. These results suggest that LLMs can generate a more diverse set of research dimensions than the ORKG properties, but with a lower degree of similarity. Notably, the nature of ORKG properties and research dimensions differs in their scope and focus. Common ORKG properties like "research problem", "method", "data source" and others provide valuable information about a specific paper, but they can not be used to comprehensively describe a research field as a whole. In contrast, research dimensions refer to shared properties of a research question, rather than focus on an individual paper. This difference contributes to the low mapping between paper properties and research field dimensions, which further consolidates our conjecture that an LLM based on only its own knowledge applied on a given research problem might not be able to completely emulate a human expert's subjectivity in defining ORKG properties. These results therefore are not a direct reflection of the inability of the LLMs tested to recommend suitable properties to structure contributions on the theme. This then opens the avenue for future work to explore how fine-tuned LLMs on the scientific domain perform on the task as a direct extension of our work.

\subsection{Embeddings-based evaluations}

The embeddings-based evaluations provide a more nuanced perspective on the semantic relationships between the ORKG properties and the LLM-generated research dimensions. By leveraging the SciNCL embeddings, we were able to quantify the cosine similarity between these two concepts, offering insights into their alignment. The results indicate a high degree of semantic similarity, with the cosine similarity between ORKG properties and the LLM-generated dimensions reaching 0.84 for GPT-3.5, 0.79 for Mistral, and 0.76 for Llama 2. These values suggest that the LLM-generated dimensions exhibit a strong correlation with the manually curated ORKG properties, signifying a substantial semantic overlap between the two.

\begin{figure}[!htb]
\includegraphics[width=10.5 cm]{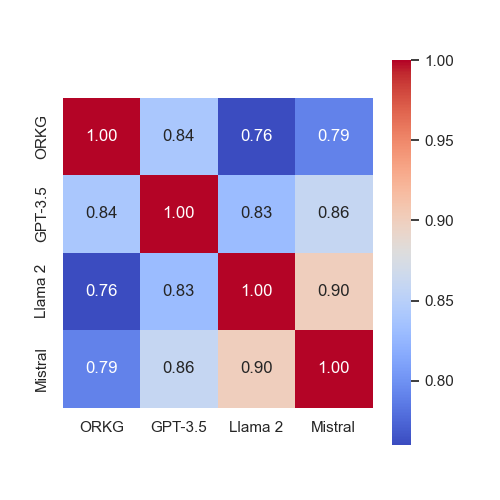}
\caption{Correlation heatmap of cosine similarity between ORKG properties and LLM-generated dimensions\label{figure Correlation Heatmap}}
\end{figure}

Furthermore, the correlation heatmap (\ref{figure Correlation Heatmap}) provides a visual representation of these similarities, highlighting the strongest correlations between ORKG properties and LLM-generated dimensions. The embeddings-based evaluations indicate that GPT-3.5 demonstrates the highest similarity to the ORKG properties, outperforming both Llama 2 and Mistral. When comparing LLM-generated dimensions between each other, a strong similarity observed between the Llama 2 and Mistral dimensions highlights the remarkable consistency in the research dimensions generated by these two models. 

Overall, the embeddings-based evaluations provide a quantitative representation of the semantic relationships between the ORKG properties and the LLM-generated research dimensions. These results suggest that while there are notable differences between the two, the LLMs exhibit a strong capacity to generate dimensions that are semantically aligned with the manually curated ORKG properties, particularly in the case of GPT-3.5. This finding highlights the potential of LLMs to serve as valuable tools for automated research metadata creation and the retrieval of related work.

\subsection{Human assessment survey}

To further evaluate the utility of the LLM-generated research dimensions, we conducted a survey to solicit feedback from domain experts who annotated the properties to create structured science summary representations or structured contribution descriptions in the ORKG. The survey was designed to assess the participants' impressions when presented with their original ORKG properties alongside the research dimensions generated by GPT-3.5.

\begin{figure}[!htb]
\includegraphics[width=\textwidth]{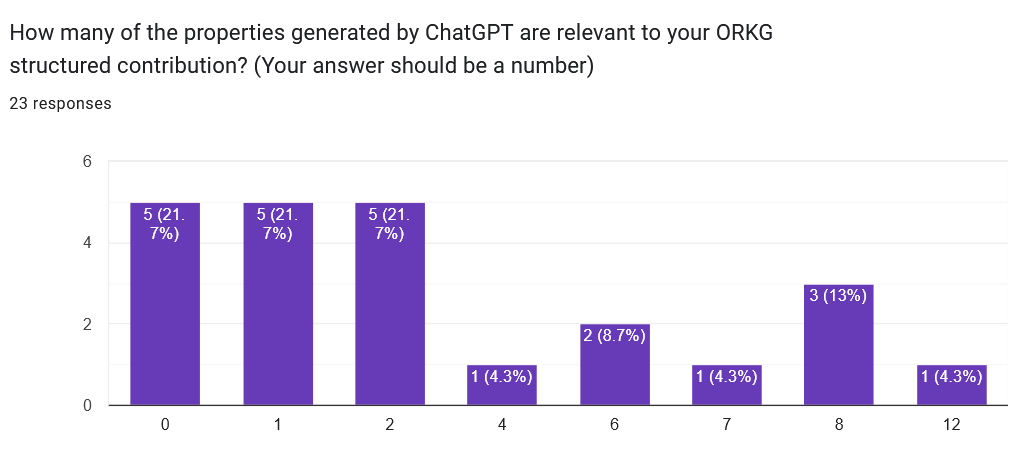}
\caption{Count of the number of relevant LLM-generated research dimensions to structure a paper's contribution or create a structured science summary.\label{figure s1q1}}
\end{figure}

\begin{figure}[!htb]
\includegraphics[width=\textwidth]{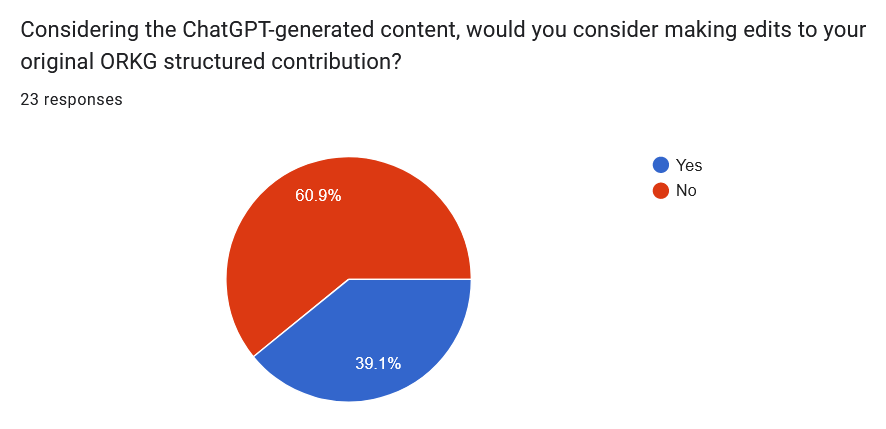}
\caption{Willingness of the participants to make changes to their existing annotated ORKG properties when shown the LLM-generated research dimensions. \label{figure s1q2}}
\end{figure}

\begin{figure}[!htb]
\includegraphics[width=\textwidth]{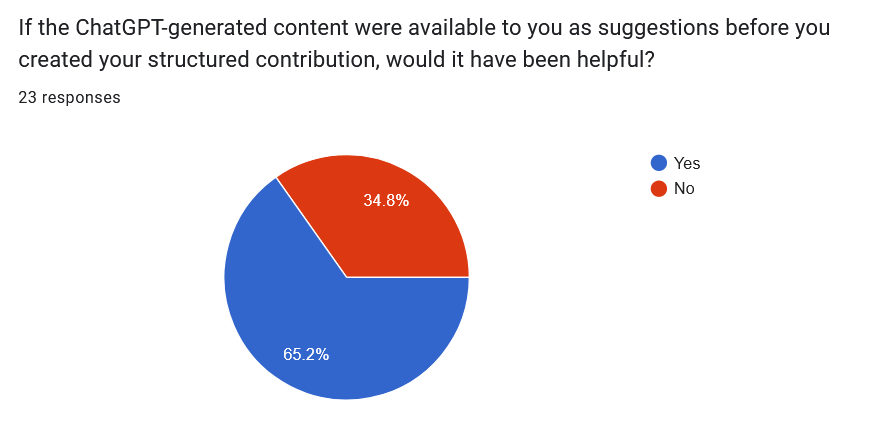}
\caption{Utility of LLM-suggested dimensions as suggestions\label{figure s1q3}}
\end{figure}

\begin{figure}[!htb]
\includegraphics[width=\textwidth]{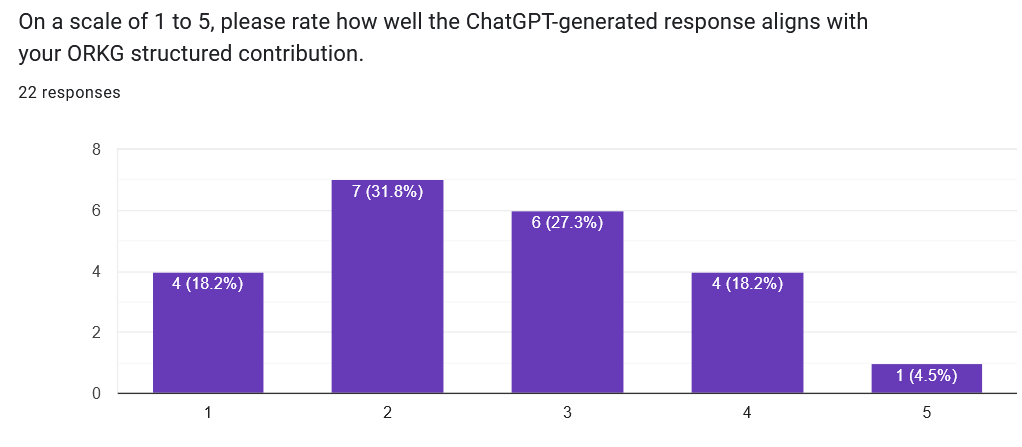}
\caption{Alignment of LLM-generated dimensions to ORKG structured contributions\label{figure s2q5}}
\end{figure}

For the first question in the Questionnaire assessing the relevance of LLM-generated dimensions to create a structured paper summary or to structure the contribution of a paper, on average, 36.3\% of the research dimensions generated by LLMs were considered highly relevant  (\autoref{figure s1q1}). This suggests that LLM-generated dimensions can provide useful suggestions for creating structured contributions on ORKG. For the second question, majority of participants (60.9\%) did not think it was necessary to make changes to their existing ORKG structure paper property annotations based on the LLM-generated dimensions, indicating that while the suggestions were relevant, they may not have been sufficient to warrant significant changes (\autoref{figure s1q2}). However, based on the third question, the survey also revealed that the majority of authors (65.2\%) believed that having LLM-generated content as suggestions before creating the structured science summaries or structured contributions would have been helpful (\autoref{figure s1q3}). The respondents noted that such LLM-based dimension suggestions could serve as a useful starting point, provide a basis for improvement, and aid in including additional properties. Based on the fourth question, the alignment between LLM-generated research dimensions and the original ORKG structured contribution properties was rated as moderate, with an average rating of 2.65 out of 5 (\autoref{figure s2q5}). This indicates that while there is some similarity between the two, there is room for alignment. As such participants raised concerns about the specificity of generated dimensions potentially diverging from the actual goal of the paper. For the final question on the release of such an LLM-based feature, the respondents emphasized the importance of aligning LLMs based on the domain expert property names while allowing descriptions to be generated, ensuring relevance across different research domains and capturing specific details like measurement values and units.

Overall, the findings of the survey indicate that LLM-generated dimensions exhibited a moderate alignment with manually extracted properties. Although the generated properties did not perfectly align with the original contributions, they still provided valuable suggestions that authors found potentially helpful in various aspects of creating structured contributions for ORKG. For instance, the suggestions were deemed useful in facilitating the creation of comparisons, identifying relevant properties, and providing a starting point for further refinement. However, concerns regarding the specificity and alignment of the generated properties with research goals were noted, suggesting areas for further refinement. These concerns highlight the need for LLMs to better capture the nuances of research goals and objectives in order to generate more targeted and relevant suggestions. Nonetheless, the overall positive feedback from participants suggests that AI tools, such as LLMs, hold promise in assisting users in creating structured research contributions and comparisons within the ORKG platform.

\section{Conclusions}

In this study, we investigated the performance of state-of-the-art Large Language Models (LLMs) in recommending research dimensions, aiming to address the central research question: How effectively do LLMs perform on the task of recommending research dimensions? Through a series of evaluations, including semantic alignment and deviation assessments, property and research dimension mappings, embeddings-based evaluations, and a human assessment survey, we sought to provide insights into the capabilities and limitations of LLMs in this domain.

The findings of our study elucidated several key aspects of LLM performance in recommending research dimensions. First, our semantic alignment and deviation assessments revealed a moderate level of alignment between manually curated ORKG properties and LLM-generated research dimensions, accompanied by a higher degree of deviation. While LLMs demonstrate some capacity to capture semantic similarities, there are notable differences between the concepts of structured paper properties and research dimensions. This suggests that LLMs may not fully emulate the nuanced inclinations of domain experts when structuring contributions.

Second, our property and research dimension mappings analysis indicated a low number of mappings between paper properties and research dimensions. While LLMs can generate a more diverse set of research dimensions than the ORKG properties, the degree of similarity is lower, highlighting the challenges in aligning LLM-generated dimensions with human expert curated properties.

Third, our embeddings-based evaluations showed that GPT-3.5 achieved the highest semantic similarity between ORKG properties and LLM-generated research dimensions, outperforming Mistral and Llama 2 in that order.

Fourth and finally, our human assessment survey provided valuable feedback from domain experts, indicating a moderate alignment between LLM-generated dimensions and manually annotated properties. While the suggestions provided by LLMs were deemed potentially helpful in various aspects of creating structured contributions, concerns regarding specificity and alignment with research goals were noted, suggesting areas for improvement.

In conclusion, our study contributes to a deeper understanding of LLM performance in recommending research dimensions to create structured science summary representations in the ORKG. While LLMs show promise as tools for automated research metadata creation and the retrieval of related work, further development is necessary to enhance their accuracy and relevance in this domain. Future research may explore the fine-tuning of LLMs on scientific domains to improve their performance in recommending research dimensions.


\section*{CRediT Author Contributions}
Author contributions are as follows: Conceptualization, Jennifer D'Souza and Steffen Eger; methodology, Vladyslav Nechakhin; validation, Vladyslav Nechakhin; investigation, Vladyslav Nechakhin and Jennifer D'Souza; resources, Vladyslav Nechakhin and Jennifer D'Souza; data curation, Vladyslav Nechakhin; writing---original draft preparation, Vladyslav Nechakhin and Jennifer D'Souza.; writing---review and editing, Jennifer D'Souza and Steffen Eger; visualization, Vladyslav Nechakhin; supervision, Jennifer D'Souza and Steffen Eger; project administration, Jennifer D'Souza; funding acquisition, Jennifer D'Souza. All authors have read and agreed to the published version of the manuscript.

\section*{Funding}

This work was supported by the German BMBF project SCINEXT (ID 01lS22070), the European Research Council for ScienceGRAPH (GA ID: 819536), and German DFG for NFDI4DataScience (no. 460234259).


\section*{Data Availability}

The evaluation dataset created in this paper is publicly accessible at \url{ https://doi.org/10.25835/6oyn9d1n}.


\section*{Acknowledgments}

We thank all members of the ORKG Team for their dedication in creating and maintaining the ORKG platform. Furthermore, we thank all the participants of our survey for providing their insightful feedback and responses.

%
%
\bibliographystyle{splncs04}
\bibliography{bibliography}

\end{document}